\documentclass{llncs}
\usepackage{amsmath}
\usepackage{amssymb}
\usepackage{times}
\usepackage{indentfirst}
\usepackage{graphics}
\usepackage{graphicx}
\usepackage{subfig}
\newtheorem{prop}{Proposition}
\newtheorem{defn}{Definition}
\newtheorem{theo}{Theorem}

\newtheorem{exa}{Example}
\newtheorem{cor}{Corollary}

\begin{document}

\title{Closed-set lattice of regular sets based on a serial and transitive relation through matroids}

\author{Qingyin Li and William Zhu\thanks{Corresponding author.
E-mail: williamfengzhu@gmail.com (William Zhu)}}
\institute{
Lab of Granular Computing,\\
Minnan Normal University, Zhangzhou 363000, China}



\date{\today}          
\maketitle

\begin{abstract}
Rough sets are efficient for data pre-processing in data mining. Matroids are based on linear algebra and graph theory, and have a
variety of applications in many fields. Both rough sets and matroids are closely related to lattices.
For a serial and transitive relation on a universe, the collection of all the regular sets of the generalized rough set is a lattice.
In this paper, we use the lattice to construct a matroid and then study relationships between the lattice and the closed-set lattice of
the matroid. First, the collection of all the regular sets based on a serial and transitive relation is proved to be a semimodular lattice.
Then, a matroid is constructed through the height function of the semimodular lattice.
Finally, we propose an approach to obtain all the closed sets of the matroid from the semimodular lattice.
Borrowing from matroids, results show that lattice theory provides an interesting view to investigate rough sets.

\end{abstract}


\section{Introduction}
Rough set theory was introduced by Pawlak \cite{Pawlak82Rough} in 1982.
It is a new mathematical tool to handle inexact, uncertain or vague knowledge and has been successfully applied to many
fields such as machine learning, pattern recognition and data mining \cite{DaiXu13Attribute,DaiXuWangTian12Conditional,InuiguchiHiranoTsumoto03RoughSet,LinYaoZadeh01Rough,Pawlak91Rough}.
In order to meet various practical applications, rough set theory has been connected with other theories, such as fuzzy set theory \cite{Liu08Generalized,Pei05AGeneralized}, boolean algebra \cite{PawlakSkowron07RoughSetAndBooleanReasoning,QiLiu05Rough}, topology \cite{LashinKozaeKhadraMedhat05Rough,WangZhuZhu10Structure,Zhu07Topological}, lattice theory \cite{ChenZhangYeungTsang06Rough,Dai08Rough,Dai05Logic,DaiChenPan06Roughsets,EstajiHooshmandaslDawaz12Rough,GhanimMustafaAziz11Onlower} and so on.

Matroid theory also has been used to study rough set theory in recent years \cite{WangZhuMin11Transversal,ZhuWang11Matroidal}.
Matroid theory \cite{Lai2001Matroidtheory} proposed by Whitney is a generalization of linear algebra, graph theory, transcendence theory and semimodular lattice theory. Matroids have been applied to a number of fields, such as combinatorial optimization \cite{Lawler01Combinatorialoptimization}, algorithm design \cite{Edmonds71Matroids}, information coding \cite{RouayhebSprintsonGeorghiades10Ontheindex} and so on.

Both rough sets and matroids are closely linked to lattices.
The fixed points of the lower approximation of the upper approximation of generalized rough sets based on relations are regular sets.
For a serial and transitive relation, the collection of all the regular sets based on the relation is a lattice.
In this paper, a matroidal structure is constructed by the lattice, and then relationships between the lattice
and the closed-set lattice of the matroid are studied. In fact, the collection of all the regular sets based on
a serial and transitive is a semimodular lattice. We define a family of subsets by the height function of the
semimodular lattice, and then prove that this family satisfies the independent set axiom of matroids. Therefore,
we obtain a matroid with the family as its independent sets. The rank function of the matroid is also represented by
the height function of the semimodular lattice. Moreover, we propose an approach to obtain all the closed sets of the
matroid from the semimodular lattice. This approach has three steps. Firstly, any singleton which is a subset of the universe
but not a subset of any atom of the semimodular lattice is a closed set of the matroid. Secondly, any element in the semimodular lattice
is a closed set of the matroid. Thirdly, if there exist two regular sets in the semimodular lattice with the property that one covers the other one and a subset of the universe whose elements are between the two ones of these two regular sets, then the closure of the subset is equal to
the regular set whose height function is one more than the remaining one.

The rest of this paper is organized as follows. In Section~\ref{S:SectionBasic},
we review some basic knowledge about generalized rough sets, matroids and lattices.
Section~\ref{S:SectionMatroidalStructure} constructs a matroidal structure by the semimodular lattice of
regular sets based on a serial and transitive relation. In Section~\ref{S:SectionRelationship}, we study
relationships between the semimodular lattice and the closed-set lattice of the matroid.
Finally, we conclude this paper in Section~\ref{S:SectionConclusion}.
\section{Basic definitions}
\label{S:SectionBasic}
In this section, we recall some basic definitions and related results of generalized rough sets, matroids and lattices.

\subsection{Generalized rough sets based on relations}

For any $x\in U$, we call $\{y\in U|xRy\}$ the successor neighborhood of $x$ in $R$ and denote it as $R_{s}(x)$.

A relation $R\subseteq U\times U$ is serial if for any $x\in U$, there exists $y\in U$ such that $xRy$,
$R$ is transitive if $xRy$ and $yRz$ imply $xRz$ for all $x$, $y$, $z \in U$.

A set $U$ with a binary relation $R$ is called a generalized approximation space.
Lower and upper approximations are two key notions in generalized approximation spaces.
In the following definition, we introduce the lower and upper approximations
of generalized approximation spaces through the successor neighborhood.

\begin{defn} (Lower and upper approximations \cite{Yao98Constructive})
 Let $(U$, $R)$ be a generalized approximation space. For any $X \subseteq U$,
\begin{center}
$\underline{R}(X)=\{x\in U|R_{s}(x)\subseteq X\}$,\\
$ \overline{R}(X)=\{x\in U|R_{s}(x)\cap X\neq \emptyset\}$,
\end{center}
are called the lower and upper approximations of $X$ in $(U$, $R)$, respectively.
\end{defn}

The lower approximation of the upper approximation operator is used to define regular sets.

\begin{defn} \cite{YangXu09Algebraicaspects}
Let $(U$, $R)$ be a generalized approximation space and $X\subseteq U$. If $X= \underline{R}\overline{R}(X)$, then $X$ is called a regular set of $(U$, $R)$. The collection of all regular sets of $(U$, $R)$ is denoted as $Reg(U$, $R)$.
\end{defn}

The collection of all regular sets based on a relation together with the set inclusion is a lattice if the relation is serial and transitive. 
For any subset of the collection of all regular sets, its least upper bound is the lower approximation of the upper approximation of the union
of all the elements in the subset and greatest lower bound is the intersection of all the elements in the subset.

\begin{prop}\label{P:Prop1} \cite{YangXu09Algebraicaspects}
Let $(U$, $R)$ be a generalized approximation space and $\{X_{i}|i\in I\}\subseteq Reg(U$, $R)$. 
If $R$ is serial and transitive, then in $(Reg(U$, $R)$, $\subseteq)$, we have\\
$(1)$ $\vee_{i\in I}X_{i}=\underline{R}\overline{R}(\cup_{i\in I}X_{i})$;\\
$(2)$ $\wedge_{i\in I}X_{i}=\cap_{i\in I}X_{i}$.
\end{prop}

The lattice defined by the collection of all regular sets based on a  serial and transitive relation is a distributive lattice.

\begin{prop} \cite{YangXu09Algebraicaspects}\label{P:PropDistributiveLattice}
Let $(U$, $R)$ be a generalized approximation space. If $R$ is serial and transitive, then  $(Reg(U$, $R)$, $\subseteq)$ is distributive.
\end{prop}

\subsection{Matroids}
A characteristic of matroids is that they can be defined in many different but equivalent ways.
In the following, a matroid is defined from the viewpoint of independent sets.

\begin{defn}(Matroid \cite{Lai2001Matroidtheory})\label{P:PropIndependentset} A matroid is a pair $(E$, $\mathbf{I})$ and it is usually denoted by $M$, where $E$ (called the ground set) is a finite set, and $\mathbf{I}$ (called the independent sets) is a family of subsets of $E$ satisfying the following three conditions:\\
$(I1)$ $\emptyset\in \mathbf{I}$;\\
$(I2)$  if $I\in \mathbf{I}$, and $I'\subseteq I$, then $I'\in \mathbf{I}$;\\
$(I3)$  if $I_{1}$, $I_{2}\in \mathbf{I}$, and $|I_{1}| < |I_{2}|$, then there exists $e\in I_{2}-I_{1}$  such that $I_{1}\cup\{e\}\in \mathbf{I}$,\\
where $|X|$ denotes the cardinality of $X$.
\end{defn}

The rank function of a matroid generalizes the maximal independence in vector subspaces.
It plays an important role in matroid theory, and it is defined as follows.

\begin{defn}(Rank function  \cite{Lai2001Matroidtheory}) Let $M = (E$, $\mathbf{I})$ be a matroid. The rank function $r_{M}$ of $M$ is defined as $r_{M}(X) = max\{|I| | I \subseteq X$, $I \in \mathbf{I}\}$ for all $X \subseteq E$. We omit the subscript $M$ when there is no confusion.
\end{defn}

\begin{prop} \cite{Lai2001Matroidtheory}
Let $M = (E$, $\mathbf{I})$ be a matroid and $r_{M}$ its rank function. For all $X \subseteq E$, $r_{M}(X) = |X|$ if and only if $X \in \mathbf{I}$.
\end{prop}

The closure operator is one of important characteristics of matroids.
A matroid and its closure operator can uniquely determine each other.
In order to represent the relationship between an element and a set of a universe,
we introduce the closure operator through the rank function in matroids.

\begin{defn} (Closure  \cite{Lai2001Matroidtheory}) Let $M = (E$, $\mathbf{I})$ be a matroid. The closure operator $cl_{M}$
of $M$ is defined as $cl_{M}(X) = \{u\in E| r_{M}(X)=r_{M}(X\cup \{u\})\}$ for all $X \subseteq E$.
$cl_{M}(X)$ is called the closure of $X$ in $M$.
\end{defn}

In a matroid, if the closure of a set is equal to itself, then the set is a closed set.
In other words, a closed set of a matroid is a fixed point of the closure operator.

\begin{defn} (Closed set  \cite{Lai2001Matroidtheory}) Let $M = (E$, $\mathbf{I})$ be a matroid and $X \subseteq E$.
We say that $X$ is a closed set of $M$ if $cl_{M}(X) = X$.
\end{defn}

\subsection{Lattices}
There are two equivalent ways to define lattices: one is based on the notion of the partially ordered set, and the other is from the viewpoint of the algebraic system. In the following definition, we will introduce one based on the notion of the partially ordered set.

\begin{defn} (Lattice \cite{Birkhoff95Lattice,Gratzer78General,Gratzer71Lattice}) \label{D:Defnlattice}
A partially ordered set $\langle L$, $\leq\rangle$ (or $L$ for short) is a lattice if every subset
$\{a$, $b\}$ of $L$ has a least upper bound $a\vee b$ and a greatest lower bound $a\wedge b$.
$L$ is said to have a least element if there exists an element $0\in L$ such that $0\vee x=x$ for all $x\in L$.
\end{defn}

\begin{defn} (Cover \cite{Birkhoff95Lattice,Gratzer78General,Gratzer71Lattice})
Let $L$ be a partially ordered set and $a$, $b\in L$. We say that $a$ is covered by $b$ (or $b$ covers $a$) and write $a\prec b$,
if $a \leq b$ and there is no element $c$ in $L$ with $a \leq c \leq b$.
\end{defn}

\begin{defn} (Atom \cite{Birkhoff95Lattice,Gratzer78General,Gratzer71Lattice})
Let $L$ be a lattice with a least element $0$. $a \in L$ is called an atom of $L$ if $a$ covers $0$. 
The set of atoms of $L$ is denoted by $\mathcal{A}(L)$.
\end{defn}

\begin{figure}[h]
\centering
\includegraphics[width=1.87cm,height=3.32cm]{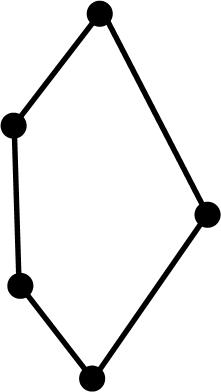}
\caption{$N_{5}$}
\label{F:figc1}
\end{figure}

\begin{theo} (Modular lattice \cite{Birkhoff95Lattice,Gratzer78General,Gratzer71Lattice})\label{T:TheoNotIsomorphicN5}
Let $L$ be a lattice. $L$ is a modular lattice iff it dose not contain a sublattice isomorphic to $N_{5}$. (See Fig.~\ref{F:figc1}).
\end{theo}

\begin{cor}\cite{ManfredStern82Rough}\label{C:Cor1}
In a modular lattice, $a \wedge  b\prec a$ if and only if $b \prec a\vee b$.
\end{cor}

\begin{defn} (Interval \cite{Birkhoff95Lattice,Gratzer78General,Gratzer71Lattice})
Let $L$ be a partially ordered set and $a$, $b\in L$ with $a\leq b$. We say that $[a$, $b]$ is an interval of $L$ if
\begin{center}
$[a$, $b]=\{c\in L|a\leq c\leq b\}$.
\end{center}
\end{defn}

If $[a$, $b]=\{a$, $b\}$, then $a$ is covered by $b$.

\begin{defn} (Linearly ordered set, chain \cite{Birkhoff95Lattice,Gratzer78General,Gratzer71Lattice})
Let $L$ be a partially ordered set. For any $a$, $b\in L$, we say that $a$ and $b$ is comparable if $a\leq b$ or $b\leq a$. $L$ is called a linearly ordered set if  $a$ and $b$ is comparable. For any $C\subseteq L$, if $C$ is a linearly ordered set, then $C$ is called a chain of $L$. The length of $C$ is $|C|-1$. A chain $C$ in $L$ is maximal if $C\cup\{x\}$ is not a chain in $L$ for any $x \in L-C$.
\end{defn}

\begin{defn}(Height function \cite{Birkhoff95Lattice,Gratzer78General,Gratzer71Lattice})
Let $L$ be a lattice with a least element $0$. For $a\in L$, then maximal chain in $[0$, $a]$ is called the height function of $a$,
and the height function of $a$ is denoted by $h(a)$.
\end{defn}

\begin{theo} \cite{Birkhoff95Lattice,Gratzer78General,Gratzer71Lattice}\label{T:TheoDistributiveModular}
Let $L$ be a lattice. If $L$ is a distributive lattice, then $L$ is also a modular lattice.
\end{theo}

\begin{defn} (Semimodular lattice \cite{Birkhoff95Lattice,Gratzer78General,Gratzer71Lattice})\label{D:DefnSemimodular}
A lattice $L$ is called a semimodular lattice if it satisfies the following condition:
\begin{center}
for any $a$, $b\in L$, $[a\wedge b$, $b]=\{a\wedge b$, $b\}$ implies that $[a$, $a\vee b]=\{a$, $a\vee b\}$.
\end{center}
\end{defn}

\section{Matroidal structure induced by regular sets based on a serial and transitive relation}
\label{S:SectionMatroidalStructure}
In this section, we propose a matroidal structure which is induced by the collection of all the regular sets
based on a serial and transitive relation. First, we prove that the collection of all the regular sets based
on a relation together with the set inclusion is a semimodular lattice if the relation is serial and transitive.

\begin{prop}\label{T:TheoSemimodular}
If $R$ is serial and transitive, then $(Reg(U$, $R)$, $\subseteq)$ is a semimodular lattice.
\end{prop}

\begin{proof}
For any $X$, $Y\in Reg(U$, $R)$, if $[X\wedge Y$, $Y]=\{X\wedge Y$, $Y\}$, then according to Definition~\ref{D:DefnSemimodular},
we need to prove that $[X$, $X\vee Y]=\{X$, $X\vee Y\}$. According to Proposition~\ref{P:PropDistributiveLattice}, Theorem~\ref{P:PropDistributiveLattice} and Corollary~\ref{C:Cor1}, it is straightforward.
\end{proof}

In the following, we define a family of subsets of the universe from the height function of the semimodular lattice.

\begin{defn}\label{D:DefnIndeprdentNew}
Let $h$ be the height function of $(Reg(U$, $R)$, $\subseteq)$. We define a family of
subsets of $U$ on $(Reg(U$, $R)$, $\subseteq)$ with respect to $h$ as follows:
\begin{center}
$\mathbf{I}(Reg(U$, $R); h)=\{X\subseteq U | h(Y)\geq|X\cap Y|$, $\forall Y\in Reg(U$, $R)\}$.
\end{center}
\end{defn}

In the following proposition, we will prove that $\mathbf{I}(Reg(U$, $R); h)$ satisfies the independent set axiom of matroids.

\begin{prop}\label{P:PropSatisfyIndependent}
  $\mathbf{I}(Reg(U$, $R); h)$ satisfies $(I1)$, $(I2)$ and $(I3)$ in Definition~\ref{P:PropIndependentset}.
\end{prop}
\begin{proof}
$(I1)$: Since $R$ is a serial relation, $R_{s}(x)\neq \emptyset$ for any $x\in U$.
Since $\emptyset\cap R_{s}(x)=\emptyset$ and $R_{s}(x)\nsubseteq \emptyset$ for any $x\in U$, $\emptyset\in Reg(U$, $R)$.
Since $h(Y)\geq |\emptyset \cap Y|=0$ for any $Y\in Reg(U$, $R)$, $\emptyset\in \mathbf{I}(Reg(U$, $R); h)$.

$(I2)$: If $I_{2}\in \mathbf{I}(Reg(U$, $R); h)$ and $I_{1}\subseteq I_{2}$, then $h(Y)\geq |I_{2} \cap Y|\geq |I_{1} \cap Y|$ for all $Y\in Reg(U$, $R)$. Hence $I_{1}\in \mathbf{I}(Reg(U$, $R); h)$.

$(I3)$: Let $I_{1}$, $I_{2} \in \mathbf{I}(Reg(U$, $R); h)$ and $|I_{1}|<|I_{2}|$.
For any $Y_{k}\in Reg(U$, $R)$, if $|I_{1}\cap Y_{k}|<h(Y_{k})$,
then $|(I_{1}\cup\{e\})\cap Y_{k}|\leq |I_{1}\cap Y_{k}|+1\leq h(Y_{k})$ for any $e\in I_{2}-I_{1}$.

For any $e\in I_{2}-I_{1}$, $h(U)\geq|I_{2} \cap U|=|I_{2}|\geq |I_{1}\cup\{e\}|=|(I_{1}\cup\{e\})\cap U|$
and $h(\emptyset)=|I_{2}\cap \emptyset|=|I_{1}\cup\{e\})\cap \emptyset|=0$.

For any $Y_{j}\in Reg(U$, $R)(\emptyset\neq Y_{j}(j\in J)\neq U)$, if $|I_{1}\cap Y_{j}|=h(Y_{j})$,
then for any $e\in I_{2}-I_{1}-\cup_{j\in J}Y_{j}\subseteq  I_{2}-I_{1}$, $|(I_{1}\cup\{e\})\cap Y_{j}|=h(Y_{j})$.
If $I_{2}-I_{1}-\cup_{j\in J}Y_{j}\neq\emptyset$, then $(I_{1}\cup\{e\})\in \mathbf{I}(Reg(U$, $R); h)$.
So we need to prove only $I_{2}-I_{1}-\cup_{j\in J}Y_{j}\neq\emptyset$.

We need to consider three cases:

Case $1$:
If there exists $j\in J$ such that $I_{2}-I_{1}-Y_{j}=\emptyset$, then $I_{2}\subseteq (I_{1}\cap I_{2})\cup Y_{j}$.
Since $|I_{1}|<|I_{2}|$, $|I_{1}-(I_{1}\cap I_{2})|<|I_{2}-(I_{1}\cap I_{2})|$. Hence $|I_{2}\cap Y_{j}|>|I_{1}\cap Y_{j}|=h(Y_{j})$,
which is contradictory to $I_{2} \in \mathbf{I}(Reg(U$, $R); h)$. So $I_{2}-I_{1}-Y_{j}\neq \emptyset$ for any $j\in J$.

Case $2$:
If there exist $j'$, $j'' \in J$ such that $I_{2}-I_{1}-Y_{j'}-Y_{j''}=\emptyset$, then there exists $a\in (I_{2}-I_{1}-Y_{j'})\cap Y_{j''}
=(I_{2}-I_{1}-Y_{j'})\cap I_{2}$, $b\in (I_{2}-I_{1}-Y_{j''})\cap Y_{j'}=(I_{2}-I_{1}-Y_{j''})\cap I_{2}$.
If $n_{1}=|(I_{2}-I_{1}-Y_{j'})\cap I_{2}|\geq 1$, $n_{2}=|(I_{2}-I_{1}-Y_{j''})\cap I_{2}|\geq 1$, $|(I_{1}\cap I_{2})\cap Y_{j'}|=n_{0}\geq 0$, $|(I_{1}\cap I_{2})\cap Y_{j''}|=n\geq 0$ and $|(I_{2}-I_{1})\cap (Y_{j'}\cap Y_{j''})|=m\geq 0$,
then $h(Y_{j''})\geq |Y_{j''}\cap I_{2}|=n+n_{1}+m$, $h(Y_{j'})\geq |Y_{j'}\cap I_{2}|=n_{0}+n_{2}+m$ and $|I_{2}|=|I_{1}\cap I_{2}|+n_{1}+n_{2}+m$.
Since $h(Y_{j'})=|I_{1}\cap Y_{j'}|=n_{0}+n_{3}\leq |I_{1}|$ and  $h(Y_{j''})=|I_{1}\cap Y_{j''}|=n+n_{4}\leq |I_{1}|$,
$n_{2}+m\leq n_{3}$ and $n_{1}+m\leq n_{4}$. Hence $|I_{2}|\leq|I_{1}\cap I_{2}|+n_{1}+n_{2}+2m\leq |I_{1}\cap I_{2}|+n_{3}+n_{4} \leq |I_{1}|$.
So $|I_{2}|\leq |I_{1}|$, which is contradictory to $|I_{1}|<|I_{2}|$.
Therefore, $I_{2}-I_{1}-Y_{j'}-Y_{j''}\neq\emptyset$ for any $j'$, $j'' \in J$.

Case $3$:
If $|I_{1}|=n$, then $J$ contains one and only one $j'$ such that $|I_{1}\cap Y_{j'}|=n$. If there exists $j_{0}\in J$ with $j_{0}\neq j'$ such that $|I_{1}\cap Y_{j_{0}}|=n$, then $|I_{1}\cap (Y_{j'}\cap Y_{j_{0}})|=n$. Since $(Y_{j'}\cap Y_{j_{0}})\in Reg(U$, $R)$,
$h(Y_{j^{'}})=|I_{1}\cap Y_{j^{'}}|=n$ and $h(Y_{j_{0}})=|I_{1}\cap Y_{j_{0}}|=n$, then $h(Y_{j'}\cap Y_{j_{0}})=n-1<|I_{1}\cap (Y_{j'}\cap Y_{j_{0}})|=n$, which is contradictory to $I_{1}\in \mathbf{I}(Reg(U$, $R); h)$. So $J$ contains one and only one
$j'$ such that $|I_{1}\cap Y_{j'}|=n$. If $|J|\geq 3$ and $\emptyset\neq Y_{j}(j\in J)\neq U$,
then $h(Y_{j''})=|I_{1}\cap Y_{j''}|<n$ for any $j''\in J-\{j'\}$.
Since $Y_{j''}\neq \emptyset$, $h(Y_{j''})\neq 0$. Since $|I_{1}\cap Y_{j'}|=n$, $I_{1}\subseteq Y_{j'}$.
If $Y_{j'}\cap Y_{j''}=\emptyset$, then $I_{1}\cap Y_{j''}=\emptyset$. Hence $h(Y_{j''})=|I_{1}\cap Y_{j''}|=0$, which is contradictory to $h(Y_{j''})\neq 0$. Therefore, $Y_{j'}\cap Y_{j''}\neq\emptyset$. If $Y_{j''}\nsubseteq Y_{j'}$, then $h(Y_{j'}\cap Y_{j''})<h(Y_{j''})$.
Since $0<h(Y_{j''})=|I_{1}\cap Y_{j''}|=m_{1}<n$ and $0\leq |(Y_{j''}-(I_{1}\cap Y_{j''}))\cap Y_{j'}|=m_{2}$, $|Y_{j'}\cap Y_{j''}|=m_{1}+m_{2}$.
So $h(Y_{j'}\cap Y_{j''})=|I_{1}\cap (Y_{j'}\cap Y_{j''})|=m_{1}$, which is contradictory to $m_{1}=h(Y_{j'}\cap Y_{j''})<h(Y_{j''})=m_{1}$.
Hence $Y_{j''}\subset Y_{j'}$. Therefore, $I_{2}-I_{1}-\cup_{j\in J}Y_{j}=I_{2}-I_{1}-Y_{j'}$.
According to Case $1$, $I_{2}-I_{1}-Y_{j'}\neq \emptyset$. Consequently, $I_{2}-I_{1}-\cup_{j\in J}Y_{j}\neq \emptyset$.
\end{proof}

Proposition~\ref{T:TheoSemimodular} and Proposition~\ref{P:PropSatisfyIndependent} show that the lattice generated by the collection of all the regular sets based on a serial and transitive relation is a semimodular lattice and this lattice can generate a matroid.

\begin{defn}\label{D:DefnMatroidNew}
The matroid with $\mathbf{I}(Reg(U$, $R); h)$ as its independent sets is denoted by $M(Reg(U$, $R))$.
We say $M(Reg(U$, $R))$ is the matroid induced by lattice $Reg(U$, $R)$.
\end{defn}

The matroid induced by the semimodular lattice can be illustrated by the following example.

\begin{figure}[h]
\centering
\includegraphics[width=3.29cm,height=3.44cm]{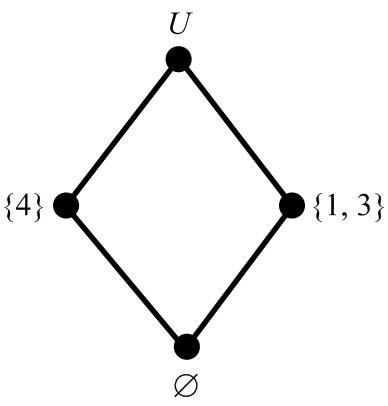}
\caption{Lattice $Reg(U$, $R)$.}
\label{F:figc2}
\end{figure}

\begin{exa}\label{E:ExaIndependentSet}
Let $U =\{1$, $2$, $3$, $4\}$ and $R = \{(1$, $1)$, $(1$, $3)$, $(2$, $1)$, $(2$, $3)$, $(2$, $4)$, $(3$, $1)$, $(3$, $3)$, $(4$, $4)\}$
a serial and transitive relation on $U$. Since $R_{s}(1) = R_{s}(3) =\{1$, $3\}$, $R_{s}(2)=\{1$, $3$, $4\}$, $R_{s}(4) =\{4\}$,
$Reg(U$, $R)=\{\emptyset$, $\{4\}$, $\{1$, $3\}$, $U\}$. The lattice $Reg(U$, $R)$ is shown in Fig.~\ref{F:figc2}. The matroid induced by $Reg(U$, $R)$ is $M(Reg(U$, $R))$ and $M(Reg(U$, $R))= (U$, $\mathbf{I}(Reg(U$, $R); h))$, where $\mathbf{I}(Reg(U$, $R); h) =
\{\emptyset$, $\{1\}$, $\{2\}$, $\{3\}$, $\{4\}$, $\{1$, $2\}$, $\{1$, $4\}$, $\{2$, $3\}$, $\{2$, $4\}$, $\{3$, $4\}\}$.
\end{exa}

In the following proposition, we obtain the expression of the rank function of the matroid
induced by the semimodular lattice $Reg(U$, $R)$ through the height function of the lattice.

\begin{prop}
Let $r$ be the rank function of $M(Reg(U$, $R))$. For all $X\subseteq U$, $r(X) = inf_{Y\in Reg(U, ~R)}\{h(Y)+|X-Y|\}$.
\end{prop}

\begin{proof}
For any $I \subseteq X$, $I \in \mathbf{I}$ and $Y\in Reg(U$,$~R)$, $h(Y)\geq |I\cap Y|$ and $|X-Y| \geq|I-Y|$, 
i.e., $h(Y)+|X-Y|\geq |I\cap Y|+|I-Y|$. So $r(X) =  max\{|I| | I \subseteq X$, $I \in \mathbf{I}\}= max\{|I\cap Y|+|I-Y| | I \subseteq X$,
$I \in \mathbf{I}$ and $Y\in Reg(U$,$~R)\}=inf_{Y\in Reg(U,~R)}\{h(Y)+|X-Y|\}$.
\end{proof}

\section{Relationships between the semimodular lattice  and the closed-set lattice of the matroid induced by the semimodular lattice}
\label{S:SectionRelationship}

In this section, we study relationships between the semimodular lattice  and the closed-set lattice of the matroid induced by the semimodular lattice.  Based on these relationships, we can obtain all the closed sets of the matroid from the semimodular lattice.

The following proposition shows that the height function of any element in the semimodular lattice is equal to
the rank function of this element with respect to the matroid induced by the lattice.

\begin{prop}\label{T:TheoRankEqualHeightFun}
Let $r$ be the rank function of $M(Reg(U$, $R))$.  For all $X\in Reg(U$, $R)$, $r(X)=h(X)$.
\end{prop}

\begin{proof}
For any $Y\in Reg(U$, $R)$, if $h(Y)\geq h(X)$ and $X\nsubseteq Y$, then there exists $Y'\in Reg(U$, $R)$ such that $h(Y)=h(Y')$ and $X\subseteq Y'$.
So $X\cap Y\subset X$ and $X\cap Y'=X$. Therefore, $\emptyset=X-Y'\subset X-Y$ and $h(Y')+|X-Y'|=h(Y')=h(Y)\geq h(X)$.
Hence, $h(Y)+|X-Y|>h(Y')+|X-Y'|\geq h(X)$.

If $h(Y)<h(X)$ and $X\cap Y=\emptyset$, then $h(Y)+|X-Y|=h(Y)+|X|\geq |X|\geq h(X)$.
If $h(Y)<h(X)$ and $X\cap Y\neq\emptyset$, then $Y\subset X$ or $Y\nsubseteq X$.
Suppose $Y\nsubseteq X$, then there exists $Y'\in Reg(U$, $R)$ such that $h(Y)=h(Y')$ and $X\cap Y\subset Y'\subset X$.
So $X\cap Y'=Y'$. If $h(Y')=h(X)-n$, $n\geq1$, then there exist $Z_{i}\in Reg(U$, $R) (i\in \{1$, $2$, $\cdots$, $n-1\})$
such that $Y'\subset Z_{n-1}\subset Z_{n-2}\subset\cdots\subset Z_{1}\subset X$
and $h(Y')=h(Z_{n-1})-1=h(Z_{n-2})-2=\cdots=h(Z_{1})-(n-1)=h(X)-n$.
So $|X-Y'|=|X|-|Y'|$ and $|Y'|\leq |Z_{n-1}|-1\leq |Z_{n-2}|-2\leq \cdots \leq |X|-n$.
Therefore, $h(Y)+|X-Y|=h(Y')+|X-Y|=h(Y')+|X-X\cap Y|>h(Y')+|X-Y'|=h(Y')+|X|-|Y'|\geq h(Y')+n = h(X)$.

Consequently, $r(X)=inf_{Y\in Reg(U,~R)}\{h(Y)+|X-Y|\}=h(X)$ for all $X\in Reg(U$, $R)$.
\end{proof}

The following proposition shows that a singleton is a closed set of the matroid induced by the semimodular lattice
if the singleton is a subset of the universe but not a subset of any atom of the lattice.

\begin{prop}\label{P:PropExcept}
Let $cl$ be the closure operator of $M(Reg(U$, $R))$. For any $e\in U-\cup \mathcal{A}(Reg(U$, $R))$, $cl(\{e\})=\{e\}$.
\end{prop}

\begin{proof}
Since $e\in U-\cup \mathcal{A}(Reg(U$, $R))$, $r(\{e\})=1$. For any $X\in \mathcal{A}(Reg(U$, $R))$ and any $x\in X$, $h(X)=r(X)=1$.
Since $h(\emptyset)+|\{x$, $e\}-\emptyset|=|\{x$, $e\}|=2$, $h(X')+|\{x$, $e\}-X'|\geq h(X')+|\{e\}|=2$ for any $X'\in\mathcal{A}(Reg(U$, $R))$
and $h(Y)+|\{x$, $e\}-Y|\geq h(Y)\geq 2$ for any $Y\in Reg(U$, $R)-\mathcal{A}(Reg(U$, $R))$, then $r(\{x$, $e\})=2$. Thus $x\notin cl(\{e\})$.
For any $y\in U-\cup \mathcal{A}(Reg(U$, $R))$ and $y\neq e$, $h(\emptyset)+|\{y$, $e\}-\emptyset|=|\{y$, $e\}|=2$.
Since $h(X)+|\{y$, $e\}-X|=h(X)+|\{y$, $e\}|=3$ for any $X\in \mathcal{A}(Reg(U$, $R))$ and $h(Y)+|\{y$, $e\}-Y|\geq h(Y)\geq 2$
for any $Y\in Reg(U$, $R)-\mathcal{A}(Reg(U$, $R))$, $r(\{y$, $e\})=2$. Hence $y\notin cl(\{e\})$. Consequently, $cl(\{e\})=\{e\}$.
\end{proof}

For the union of any regular set $X$ in the semimodular lattice and any singleton which is a subset of the complement of $X$ in the universe,
its rank function is one more than the height function of $X$.

\begin{prop}\label{P:PropLatticeElementClouse2}
Let $r$ be the rank function of $M(Reg(U$, $R))$. For any $X\in Reg(U$, $R)$ and $e\in U-X$,  $r(X\cup \{e\})=h(X)+1$.
\end{prop}

\begin{proof}
For any $Y$, $Y'\in Reg(U$, $R)$, if $h(Y)=h(Y')\geq h(X)$, $X\nsubseteq Y$ and $X\subseteq Y'$, then $\emptyset=X-Y'\subset X-Y$.
Since $X-Y\subseteq X\cup \{e\}-Y$ for any $e\in U-X$, $h(Y)+|X\cup\{e\}-Y|=h(Y')+|X\cup\{e\}-Y|\geq h(Y')+|X-Y| >h(Y')\geq h(X)$.
If $Y'=X$, then $h(Y')+|X\cup\{e\}-Y'|=h(X)+|X\cup\{e\}-X|=h(X)+1$. If $X\subset Y'$, then $h(Y')\geq h(X)+1$ and
$h(Y')+1\geq h(Y')+|X\cup\{e\}-Y'|\geq h(Y')$. So $h(Y')+|X\cup\{e\}-Y'|\geq h(X)+1$.

If $h(Y)<h(X)$ and $X\cap Y=\emptyset$, then $Y=\emptyset$ or $Y\neq \emptyset$.
If $Y\neq\emptyset$, $h(Y)+|X\cup\{e\}-Y|\geq h(Y)+|X-Y|=h(Y)+|X|> |X|\geq h(X)$.
If $Y =\emptyset$, $h(\emptyset)+|X\cup\{e\}-\emptyset|=|X|+1\geq h(X)+1$.
If $h(Y)<h(X)$ and $X\cap Y\neq \emptyset$, then $Y\subset X$ or $Y\nsubseteq X$.
Suppose $Y\nsubseteq X$, then there exists $Y'\in Reg(U$, $R)$ such that $h(Y)=h(Y')$ and $X\cap Y\subset Y'\subset X$.
Since $e\in U-X$, $e\notin Y'$. If $h(Y')=h(X)-n$, $n\geq 1$, then there exist $Z_{i}\in Reg(U$, $R) (i\in \{1$, $2$, $\cdots$, $n-1\})$
such that $Y'\subset Z_{n-1}\subset Z_{n-2}\subset\cdots\subset Z_{1}\subset X$ and $h(Y')=h(Z_{n-1})-1=h(Z_{n-2})-2=\cdots=h(Z_{1})-(n-1)=h(X)-n$.
So $|X\cup\{e\}-Y'|=|X\cup\{e\}|-|Y'|=|X|+1-|Y'|$ and $|Y'|\leq |Z_{n-1}|-1\leq |Z_{n-2}|-2\leq \cdots \leq |X|-n$.
Therefore, $h(Y)+|X\cup\{e\}-Y|=h(Y')+|X\cup\{e\}-Y|=h(Y')+|X\cup\{e\}-X\cap Y|\geq h(Y')+|X\cup\{e\}-Y'|=h(Y')+|X|+1-|Y'|\geq h(Y')+n+1 = h(X)+1$.

Consequently, $r(X\cup\{e\})=inf_{Y\in Reg(U,~ R)}\{h(Y)+|X\cup\{e\}-Y|\}=h(X)+1$ for all $X\in Reg(U$, $R)$ and $e\in U-X$.

\end{proof}

The following proposition shows that any element in the semimodular lattice is a closed set in the matroid induced by the lattice.

\begin{prop}\label{P:PropLatticeElementClouse}
Let $cl$ be the closure operator of $M(Reg(U$, $R))$. For any $X\in Reg(U$, $R)$, $cl(X)=X$.
\end{prop}

\begin{proof}
According to Proposition~\ref{T:TheoRankEqualHeightFun}, $r(X)=h(X)$. According to Proposition~\ref{P:PropLatticeElementClouse2},
$r(X\cup \{e\})=h(X)+1$ for any $X\in Reg(U$, $R)$ and $e\in U-X$. Hence $cl(X)=X$.
\end{proof}

In the following part, we use $\mathcal{F}(n)$ to represent the set in which the height function of any element is equal to $n$.
Hence for the lattice $Reg(U$, $R)$, $\mathcal{F}(1)= \mathcal{A}(Reg(U$, $R))$.

For any two regular sets $X$ and $Y$ in the semimodular lattice, if $X$ covers $Y$,
then a set $Z$ with $Y\subset Z\subset X$ has the same height function as $X$.

\begin{prop}\label{P:PropRankClouse}
Let $r$ be the rank function of $M(Reg(U$, $R))$, $X$, $Y\in Reg(U$, $R)$ and $X\in \mathcal{F}(n)$, $Y\in \mathcal{F}(n-1)(1\leq n\leq h(U))$.
For any $Z\subseteq U$, if $Y\subset Z\subset X$, then $r(Z)=h(X)$.
\end{prop}

\begin{proof}
For any $X'\in Reg(U$, $R)$, if $h(X')\geq h(X)$ and $X\nsubseteq X'$, then there exists $Y'\in Reg(U$, $R)$ such that $h(X')=h(Y')$
and $X\subseteq Y'$. Since $Z\subset X$, $\emptyset=Z-Y'$ and $\emptyset\subseteq Z-X'$. Moreover, $h(Y')+|Z-Y'|=h(Y')\geq h(X)$.
Therefore, $h(X')+|Z-X'|=h(Y')+|Z-X'|\geq h(Y')\geq h(X)$.

If $h(X')<h(X)$ and $X'\nsubseteq X$, then there exists $Y'\in Reg(U$, $R)$ such that $h(X')=h(Y')$ and $Y'\subset Z$.
If $h(Y')=h(X)-m$, $m\geq1$, then there exist $Y_{i}\in Reg(U$, $R) (i\in \{1$, $2$, $\cdots$, $m-2\})$
such that $h(Y')=h(Y_{m-2})-1=h(Y_{m-3})-2=\cdots=h(Y)-(m-1)=h(X)-m$ and $Y'\subset Y_{m-1}\subset\cdots\subset Y_{1}\subset Y \subset  X$.
Since $|Z|-|Y| \geq 1$, $|Y|-|Y_{1}| \geq 1$, $\cdots $, $|Y_{m-2}|-|Y'|\geq 1$, $|Z|-|Y|+|Y|-|Y_{1}|+\cdots+|Y_{m-2}|-|Y'|\geq m$.
So $|Z-Y'|=|Z|-|Y'|\geq m$. Therefore, $h(Y')+|Z-Y'|=h(Y')+|Z|-|Y'|\geq h(Y')+m= h(X)$.
Hence $h(X')+|Z-X'|=h(Y')+|Z-X'|>h(Y')+|Z-Y'|\geq h(X)$.
If $X'\subset X$ but $X'\nsubseteq Y$, then $|Z-X'|>|Z-Y'|$.
So we also have $h(X')+|Z-X'|=h(Y')+|Z-X'|>h(Y')+|Z-Y'|\geq h(X)$.

Therefore, $r(Z)=inf_{Y\in Reg(U, ~R)}\{h(Y)+|Z-Y|\}=h(X)$.
\end{proof}

For any two regular sets $X$ and $Y$ in the semimodular lattice and any subset $Z$ of the universe, if $Y$ is covered by $X$ and $Y\subset Z\subset X$, then the closure of $Z$ with respect to the matroid induced by the lattice is equal to $X$.

\begin{prop}\label{C:CorTwoLayerSubset}
Let $cl$ be the closure operator of $M(Reg(U$, $R))$, $X$, $Y\in Reg(U$, $R)$ and $X\in \mathcal{F}(n)$, $Y\in \mathcal{F}(n-1)(1\leq n\leq h(U))$. For any $Z\subseteq U$, if $Y\subset Z\subset X$,  then $cl(Z)=X$.
\end{prop}

\begin{proof}
Since $X\in \mathcal{F}(n)$, $Y\in \mathcal{F}(n-1)$ and $Y\subset Z\subset X$, then according to Proposition~\ref{P:PropRankClouse}, $r(Z)=h(X)$.
According to Proposition~\ref{T:TheoRankEqualHeightFun}, $r(X)=h(X)$. According to Proposition~\ref{P:PropLatticeElementClouse}, $cl(X)=X$.
Therefore, $cl(Z)=X$.
\end{proof}

Given a matroid $M$, it is well known the family of all the closed sets of the matroid together with the set inclusion is a closed-set lattice \cite{Lai2001Matroidtheory}. The closed-set lattice of $M$ is denoted by $\mathcal{L}(M)$. For any $X$, $Y\in \mathcal{L}(M)$, their least upper bound and greatest lower bound are $X\vee Y=cl(X\cup Y)$ and $X\wedge Y=X\cap Y$, respectively, where $cl$ is the closure operator of $M$.
The following corollary shows the height function of any element in the semimodular lattice is equal to the height function of itself within the closed-set lattice induced by the matroid which is generated by the semimodular lattice.

\begin{cor}\label{C:CorHeightEqual}
Let $h_{1}$ be the height function of $\mathcal{L}(M(Reg(U$, $R)))$. For any $X\in Reg(U$, $R)$, $h(X)=h_{1}(X)$.
\end{cor}

\begin{proof}
Let $X$, $Y\in Reg(U$, $R)$, $X\in \mathcal{F}(n)$ and $Y\in \mathcal{F}(n-1)$.
According to Corollary~\ref{C:CorTwoLayerSubset}, $cl(Z)=X$ for any $Z\subseteq U$ with $Y\subset Z\subset X$.
Hence $Z\notin \mathcal{L}(M(Reg(U$, $R)))$. Therefore, $h(X)=h_{1}(X)$ for any  $X\in Reg(U$, $R)$.
\end{proof}

According to Proposition~\ref{P:PropExcept}, Proposition~\ref{P:PropLatticeElementClouse}, Proposition~\ref{C:CorTwoLayerSubset} and Corollary~\ref{C:CorHeightEqual}, we can obtain all the elements in $\mathcal{L}(M(Reg(U$, $R)))$ from $Reg(U$, $R)$.
In order to illustrate this feature, we present the following example.

\begin{figure}[h]
\centering
\includegraphics[width=3.29cm,height=3.47cm]{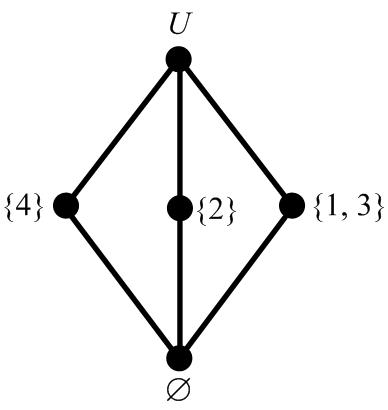}
\caption{Lattice $\mathcal{L}(M(Reg(U$, $R)))$.}
\label{F:figc3}
\end{figure}

\begin{exa}
As shown in Example~\ref{E:ExaIndependentSet}, $Reg(U$, $R)=\{\emptyset$, $\{4\}$, $\{1$, $3\}$, $U\}$.
According to Proposition~\ref{P:PropExcept}, $\{2\}\in \mathcal{L}(M(Reg(U$, $R)))$.
According to Proposition~\ref{P:PropLatticeElementClouse}, $\{\emptyset$, $\{4\}$, $\{1$, $3\}$, $U\}\subseteq \mathcal{L}(M(Reg(U$, $R)))$.
According to Proposition~\ref{C:CorTwoLayerSubset} and Corollary~\ref{C:CorHeightEqual}, $Z\notin \mathcal{L}(M(Reg(U$, $R)))$ for any $ Z\subset U$
with $\{4\}$ or $\{1$, $3\}\subset Z\subset U$ or $\emptyset\subset Z\subset\{1$, $3\}$.
Therefore, $\mathcal{L}(M(Reg(U$, $R)))=\{\emptyset$, $\{2\}$, $\{4\}$, $\{1$, $3\}$, $U\}$.
The lattice $\mathcal{L}(M(Reg(U$, $R)))$ is shown in Fig.~\ref{F:figc3}.
\end{exa}

\section{Conclusions}
\label{S:SectionConclusion}
In this paper, we constructed a matroid by the semimodular lattice of regular sets based on a serial and transitive relation,
and studied relationships between the semimodular lattice and the closed-set lattice of the matroid.
A family of subsets was defined from the semimodular lattice, which provided a platform for studying rough sets using matroidal approaches.
For example, regular sets of the generalized rough set based on a serial and transitive relation and the closure of certain subset of the universe were connected. In future, we will use lattice theory to study relationships between covering-based rough sets and matroids.

\section{Acknowledgments}
This work is supported in part by the National Natural Science Foundation
of China under Grant No. 61170128, the Natural Science Foundation of Fujian
Province, China, under Grant Nos. 2011J01374 and 2012J01294, and the Science
and Technology Key Project of Fujian Province, China, under Grant No.
2012H0043.

\bibliographystyle{splncs}

\begin{thebibliography}{10}


\bibitem{Birkhoff95Lattice}
Birkhoff, G.:
\newblock Lattice theory.
\newblock American Mathematical Society (1995)

\bibitem{ChenZhangYeungTsang06Rough}
Chen, D., Zhang, W., Yeung, D., Tsang, E.:
\newblock Rough approximations on a complete completely distributive lattice
  with applications to generalized rough sets.
\newblock Information Sciences \textbf{176} (2006)  1829--1848

\bibitem{Dai08Rough}
Dai, J.:
\newblock Rough 3-valued algebras.
\newblock Information Sciences \textbf{178} (2008) 1986-1996


\bibitem{Dai05Logic}
Dai, J.:
\newblock Logic for Rough Sets with Rough Double Stone Algebraic Semantics.
\newblock In: Proceedings of the 10th International Conference on Rough Sets, Fuzzy Sets, Data Mining, and Granular Computing. (2005)  141-148


\bibitem{DaiChenPan06Roughsets}
Dai, J., Chen, W., Pan, Y.:
\newblock Rough Sets and Brouwer-Zadeh Lattices.
\newblock In: Proceedings of the 1st International Conference on Rough Sets and Knowledge Technology. (2006)  200-207



\bibitem{DaiXu13Attribute}
Dai, J., Xu, Q.:
\newblock Attribute selection based on information gain ratio in fuzzy rough set theory with application to tumor classification.
\newblock Applied Soft Computing \textbf{13} (2013) 211-221

\bibitem{DaiXuWangTian12Conditional}
Dai, J., Xu, Q., Wang, W., Tian, H.:
\newblock Conditional entropy for incomplete decision systems and its application in data mining.
\newblock International Journal of General Systems \textbf{41} (2012) 713-728

\bibitem{Edmonds71Matroids}
Edmonds, J.:
\newblock Matroids and the greedy algorithm.
\newblock Mathematical Programming \textbf{1} (1971)  127--136

\bibitem{EstajiHooshmandaslDawaz12Rough}
Estaji, A., Hooshmandasl, M., Davvaz, B.:
\newblock Rough set theory applied to lattice theory.
\newblock Information Sciences \textbf{200} (2012)  108--122



\bibitem{GhanimMustafaAziz11Onlower}
Ghanim, M., Mustafa, H., Aziz, S.:
\newblock On lower and upper intension order relations by different cover
  concepts.
\newblock Information Sciences \textbf{181} (2011)  3723--3734

\bibitem{Gratzer78General}
Gratzer, G.:
\newblock General lattice theory.
\newblock New York, San Francisco: Academic Press (1978)

\bibitem{Gratzer71Lattice}
Gratzer, G.:
\newblock Lattice Theory: First Concepts and Distributive Lattices.
\newblock San Francisco: W. H. Freeman and Company (1971)


\bibitem{InuiguchiHiranoTsumoto03RoughSet}
Inuiguchi, M., Hirano, S., Tsumoto(Eds.), S.:
\newblock Rough Set Theory and Granular Computing. Volume 125.
\newblock Studies in Fuzziness and Soft Computing,Springer-Verlag, Heidelberg
  (2003)

\bibitem{Lai2001Matroidtheory}
Lai, H.:
\newblock Matroid theory.
\newblock Higher Education Press, Beijing (2001)

\bibitem{LashinKozaeKhadraMedhat05Rough}
Lashin, E., Kozae, A., Khadra, A., Medhat, T.:
\newblock Rough set theory for topological spaces.
\newblock International Journal of Approximate Reasoning \textbf{40} (2005)
  35--43


\bibitem{Lawler01Combinatorialoptimization}
Lawler, E.:
\newblock Combinatorial optimization: networks and matroids.
\newblock Dover Publications (2001)

\bibitem{LinYaoZadeh01Rough}
Lin, T., Yao, Y., Zadeh(Eds.), L.:
\newblock Rough sets, Granular Computing and Data Mining.
\newblock Studies in Fuzziness and Soft Computing, Physica-Verlag, Heidelberg,
  (2001)

\bibitem{Liu08Generalized}
Liu, G.:
\newblock Generalized rough sets over fuzzy lattices.
\newblock Information Sciences \textbf{178} (2008)  1651--1662


\bibitem{Pawlak82Rough}
Pawlak, Z.:
\newblock Rough sets.
\newblock International Journal of Computer and Information Sciences
  \textbf{11} (1982)  341--356

\bibitem{Pawlak91Rough}
Pawlak, Z.:
\newblock Rough sets: theoretical aspects of reasoning about data.
\newblock Kluwer Academic Publishers, Boston (1991)

\bibitem{PawlakSkowron07RoughSetAndBooleanReasoning}
Pawlak, Z., Skowron, A.:
\newblock Rough sets and boolean reasoning.
\newblock Information Sciences \textbf{177} (2007)  41--73

\bibitem{Pei05AGeneralized}
Pei, D.:
\newblock A generalized model of fuzzy rough sets.
\newblock International Journal of General Systems \textbf{34} (2005)  603--613



\bibitem{QiLiu05Rough}
Qi, G., Liu, W.:
\newblock Rough operations on boolean algebras.
\newblock Information Sciences \textbf{173} (2005)  49--63


\bibitem{RouayhebSprintsonGeorghiades10Ontheindex}
Rouayheb, S., Sprintson, A., Georghiades, C.:
\newblock On the index coding problem and its relation to network coding and
  matroid theory.
\newblock IEEE transactions on information theory \textbf{56} (2010)
  3187--3195

\bibitem{ManfredStern82Rough}
Stern, M.:
\newblock Semimodular lattices: theory and applications.
\newblock  Encyclopedia of Mathematics and its Applications (1999)

\bibitem{WangZhuZhu10Structure}
Wang, S., Zhu, P., Zhu, W.:
\newblock Structure of covering-based rough sets.
\newblock International Journal of Mathematical and Computer Sciences
  \textbf{6} (2010)  147--150

\bibitem{WangZhuMin11Transversal}
Wang, S., Zhu, W., Min, F.:
\newblock Transversal and function matroidal structures of covering-based rough sets.
\newblock In: Rough Sets and Knowledge Technology. (2011)  146--155


\bibitem{YangXu09Algebraicaspects}
Yang, L., Xu, L.:
\newblock Algebraic aspects of generalized approximation spaces.
\newblock International Journal of Approximate Reasoning \textbf{51} (2009)
  151--161

\bibitem{Yao98Constructive}
Yao, Y.:
\newblock Constructive and algebraic methods of theory of rough sets.
\newblock Information Sciences \textbf{109} (1998)  21--47



\bibitem{Zhu07Topological}
Zhu, W.:
\newblock Topological approaches to covering rough sets.
\newblock Information Sciences \textbf{177} (2007)  1499--1508



\bibitem{ZhuWang11Matroidal}
Zhu, W., Wang, S.:
\newblock Matroidal approaches to generalized rough sets based on relations.
\newblock International Journal of Machine Learning and Cybernetics \textbf{2}
  (2011)  273--279

\end{thebibliography}

\end{document}